\documentclass[9pt,twocolumn,twoside]{pnas-new}

\templatetype{pnasresearcharticle}
\setboolean{displaywatermark}{false}

\usepackage{amsmath, amsfonts, dsfont, mathtools}
\usepackage{xspace}

\definecolor{dark-red}{rgb}{0.8, 0.0, 0.1803921568627451}
\definecolor{dark-blue}{rgb}{0.0, 0.0, 0.803921568627451}
\definecolor{dark-green}{rgb}{0.0, 0.39215686274509803, 0.0}
\definecolor{dark-orange}{rgb}{0.8, 0.4, 0.0}

\newcommand{\diff}{\mathrm{d}}
\newcommand{\eg}{{e.\,g.}~}
\newcommand{\ie}{{i.\,e.}~}
\newcommand{\iid}{{i.i.d.}}

\newcommand{\abc}{\textsc{ABC}\xspace}

\newcommand{\intz}{\int \! \diff z\;}

\setlist[itemize]{itemsep=1pt,parsep=1pt,topsep=1.5pt}
\setlist[enumerate]{itemsep=1pt,parsep=1pt,topsep=1.5pt}

\title{The frontier of simulation-based inference}

\author[a,b,1]{Kyle Cranmer}
\author[a,b]{Johann Brehmer}
\author[c]{Gilles Louppe}

\affil[a]{Center for Cosmology and Particle Physics, New York University, USA}
\affil[b]{Center for Data Science, New York University, USA}
\affil[c]{Montefiore Institute, University of Li\`ege, Belgium}

\leadauthor{Cranmer}

\significancestatement{We provide an overview of how recent developments in machine learning are revolutionizing simulation-based inference and the profound impact this has on the physical sciences.}

\authorcontributions{Author contributions: K.\,C., J.\,B., and G.\,L.\ wrote the paper.}
\authordeclaration{The authors declare no conflict of interest.}
\correspondingauthor{\textsuperscript{1}To whom correspondence should be addressed. E-mail: kyle.cranmer@nyu.edu}

\keywords{Statistical inference $|$ Implicit models $|$ Likelihood-free inference $|$ Approximate Bayesian Computation $|$ Neural density estimation}

\begin{abstract}
Many domains of science have developed complex simulations to describe phenomena of interest. While these simulations provide high-fidelity models, they are poorly suited for inference and lead to challenging inverse problems. We review the rapidly developing field of simulation-based inference and identify the forces giving new momentum to the field. Finally, we describe how the frontier is expanding so that a broad audience can appreciate the profound change these developments may have on science.
\end{abstract}

\dates{\today}

\begin{document}

\maketitle
\thispagestyle{firststyle}
\ifthenelse{\boolean{shortarticle}}{\ifthenelse{\boolean{singlecolumn}}{\abscontentformatted}{\abscontent}}{}

\begin{figure*}
    \centering
    \includegraphics[width=0.9\textwidth]{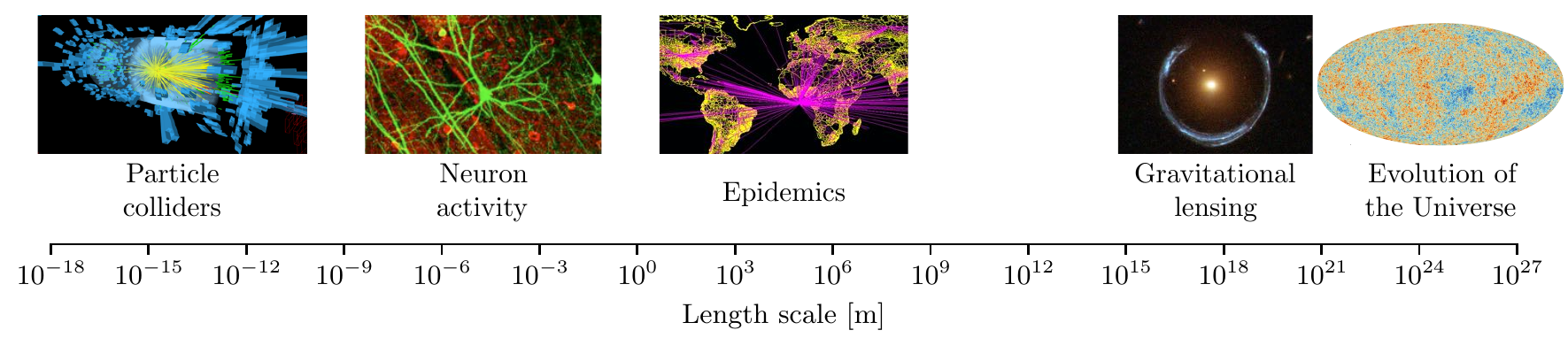}
    \caption{Examples of phenomena at various length scales described by a diverse set of simulators, each with an intractable likelihood. Contains image material from Refs.~\cite{CMS:2021490, 10.1371/journal.pbio.0040029, epidemiology-image, hubble-image, planck-image}.}
    \label{fig:simulators}
\end{figure*}

\dropcap{M}echanistic models can be used to predict how systems will behave in a variety of circumstances. These run the gamut of distance scales with notable examples including particle physics, molecular dynamics, protein folding, population genetics, neuroscience, epidemiology, economics, ecology, climate science, astrophysics, and cosmology (see Fig.~\ref{fig:simulators}). The expressiveness of programming languages facilitates the development of complex, high-fidelity simulations and the power of modern computing provides the ability to generate synthetic data from them. Unfortunately, these simulators are poorly suited for statistical inference. The source of the challenge is that the probability density (or likelihood) for a given observation---an essential ingredient to both frequentist and Bayesian inference methods---is typically intractable. Such models are often referred to as implicit models and contrasted against prescribed models where the likelihood for an observation can be explicitly calculated~\cite{Diggle1984MonteCM}. The problem setting of statistical inference under intractable likelihoods has been dubbed likelihood-free inference---though it is a bit of a misnomer as typically one attempts to estimate the intractable likelihood, so we feel the term simulation-based inference is more apt.

The intractability of the likelihood is an obstruction for scientific progress as statistical inference is a key component of the scientific method. In areas where this obstruction has appeared, scientists have developed various ad-hoc or field-specific methods to overcome it. In particular, two common traditional approaches rely on scientists to use their insight into the system to construct powerful summary statistics and then compare the observed data to the simulated data. In the first, density estimation methods are used to approximate the distribution of the summary statistics from samples generated by the simulator~\cite{Diggle1984MonteCM}. This approach was used for the discovery of the Higgs boson in a frequentist paradigm and is illustrated  in Fig.~\ref{fig:workflows}e). Alternatively, a technique known as Approximate Bayesian Computation  (\abc)~\cite{rubin1984, beaumont2002approximate}  compares the observed and simulated data based on some distance measure involving the summary statistics. \abc is widely used in population biology, computational neuroscience, and cosmology and is depicted in Fig.~\ref{fig:workflows}a). Both techniques have served a large and diverse segment of the scientific community.

Recently, the toolbox of simulation-based inference has experienced an accelerated expansion. Broadly speaking, three forces are giving new momentum to the field. First, there has been a significant cross-pollination between those studying simulation-based inference and those studying probabilistic models in machine learning~\cite{2016arXiv161003483M}, and the impressive growth of machine learning capabilities enables new approaches. Second, active learning---the idea of continuously using the acquired knowledge to guide the simulator---is being recognized as a key idea to improve the sample efficiency of various inference methods. A third direction of research has stopped treating the simulator as a black box and focused on integrations that allow the inference engine to tap into the internal details of the simulator directly.

Amidst this ongoing technological revolution, the landscape of simulation-based inference is changing rapidly. In this review we aim to provide the reader with a high-level overview of the basic ideas behind both old and new inference techniques. Rather than discussing the algorithms in technical detail, we focus on the current frontiers of research, and comment on some ongoing developments that we deem particularly exciting.

We begin by describing simulation-based inference and the traditional approaches in Sec.~\ref{sec:sbi}. In Sec.~\ref{sec:frontiers} we discuss three main directions of technological progress. We then show how they can be combined in different workflows for simulation-based inference in Sec.~\ref{sec:inference}. We conclude with a discussion of the future of simulation-based inference in Sec.~\ref{sec:conclusions}.

\section{Simulation-based inference}
\label{sec:sbi}

\subsection{Simulators}
\label{sec:simulators}

Statistical inference is performed within the context of a statistical model, and in simulation-based inference the simulator itself defines the statistical model. For the purpose of this paper, a simulator is a computer program that takes as input a vector of parameters $\theta$, samples a series of internal states or latent variables $z_i \sim p_i(z_i|\theta, z_{<i})$, and finally produces a data vector $x \sim p(x|\theta, z)$ as output. Programs that involve random samplings and are interpreted as statistical models are known as \emph{probabilistic programs}, and simulators are an example. Within this general formulation, real-life simulators can vary substantially:
\begin{itemize}
    \item The parameters $\theta$ describe the underlying mechanistic model and thus affect the transition probabilities $p_i(z_i|\theta, z_{<i})$. Typically the mechanistic model is interpretable by a domain scientist and $\theta$ has relatively few components and a fixed dimensionality.
    Examples include coefficients found in the Hamiltonian of a physical system, the virulence and incubation rate of a pathogen, or fundamental constants of Nature.
    \item The latent variables $z$ that appear in the data-generating process may directly or indirectly correspond to a physically meaningful state of a system, but typically this state is unobservable in practice. The structure of the latent space varies substantially  between simulators. The latent variables may be continuous or discrete and the dimensionality of the latent space may be fixed or may vary depending on the control flow of the simulator. The simulation can freely combine deterministic and stochastic steps. The deterministic components of the simulator may be differentiable or may involve discontinuous control flow elements. In practice, some simulators may provide convenient access to the latent variables, while others are effectively black boxes. Any given simulator may combine these different aspects in almost any way.
    \item Finally, the output data $x$ correspond to the observations. They can range from a few unstructured numbers to high-dimensional and highly structured data, such as images or geospatial information.
\end{itemize}

Consider for instance the systems shown in Fig.~\ref{fig:simulators}. Particle physics processes often only depend on a small number of parameters of interest such as particle masses or coupling strengths. The latent process combines a high-energy interaction,  rigorously described by a quantum field theory, with the passage of the resulting particles through an incredibly complex detector, most accurately modeled with stochastic simulations with billions of latent variables; this second part often does not depend on the parameters of interest. The output data consist, in their raw form, of millions of sensor read-outs, though there is an established pipeline that compresses this raw data to tens to hundreds of observables. Epidemiological simulations can be based on a network structure with geospatial properties, and the latent process consists of many repeated structurally identical stochastic time steps. In contrast, cosmological simulations of the evolution of the Universe may consist of a highly structured stochastic initial state followed by a smooth, deterministic time evolution.

These differences mean that there is no one-size-fits-all inference method. In this review we aim to clarify the considerations needed to choose the most appropriate approach for a given problem.

\subsection{Inference}

Scientific inference tasks differ by \emph{what} is being inferred: given observed data $x$, is the goal to infer the input parameters $\theta$, or the latent variables $z$, or both? Sometimes only a subset of the parameters (or latent variables) are of interest, while the rest are nuisance parameters (\ie parameters that we are not directly interested in but must account for because they influence the distributions of the data). We will focus on the common problem of inferring $\theta$ in a parametric setting, we will comment on methods that allow inference on $z$, and we will not focus on non-parametric inverse problems.

Inference may be performed either in a frequentist or a Bayesian approach and may be limited to point estimates $\hat{\theta}(x)$ or extended to include a probabilistic notion of uncertainty. In the frequentist case, confidence sets are often formed from inverting hypothesis tests, based on the likelihood ratio test statistic. In Bayesian inference, the goal is typically to calculate the posterior
\begin{equation}
    p(\theta|x) = \frac{p(x|\theta) \, p(\theta)}{\int \!\! \diff \theta' \, p(x|\theta') \, p(\theta')}
    \label{eq:bayes}
\end{equation}
for observed data $x$ and a given prior $p(\theta)$. In both cases the likelihood function $p(x|\theta)$ is a key ingredient.

The fundamental challenge for simulation-based inference problems is that the likelihood function $p(x|\theta)$ implicitly defined by the simulator is typically not tractable, as it corresponds to an integral over all possible trajectories through the latent space, \ie all possible execution traces of the simulator. That is,
\begin{equation}
    p(x | \theta) = \intz p(x, z | \theta) \,,
\end{equation}
where $p(x, z | \theta)$ is the joint probability density of data $x$ and latent variables $z$. For a simple sequential data generation procedure, the joint likelihood can be written as $p(x, z | \theta) = p(x | \theta, z) \prod_{i}  p_i(z_i | \theta, z_{<i})$. For real-life simulators with large latent spaces, it is clearly impossible to compute this integral explicitly. Since the likelihood function is the central ingredient to both frequentist and Bayesian inference, this is a major challenge for inference in many fields. This paper reviews simulation-based or likelihood-free inference techniques that enable frequentist or Bayesian inference despite this intractability. These methods can be seen as a specialization of inverse Uncertainty Quantification (UQ) on the model parameters in situations with accurate, stochastic simulators.

There is a second, more widely appreciated source of intractability. In the case of Bayesian inference, the evidence---the denominator of \eqref{eq:bayes}---involves an integral over the parameters $\theta$. In problems with high-dimensional parameters this becomes intractable, independently of the intractability of the likelihood function. This challenge is commonly addressed with Markov Chain Monte Carlo (MCMC) methods~\cite{metropolis1953equation, Hastings1970} or variational inference (VI)~\cite{anderson1987mean}.

In practice, an important distinction is that between inference based on a single observation, and that based on multiple independent and identically distributed (\iid) observations. In the second case, the likelihood factorizes into individual likelihood terms for each \iid~observation, as $p(x|\theta) = \prod_i p_\mathrm{individual}(x_i | \theta)$. For example, time-series data is typically non-\iid~and must be treated as a single high-dimensional observation, whereas the analysis of collision data in the search for the Higgs boson constitutes a data set with many \iid~measurements. This distinction is important when it comes to the computational cost of an inference technique, as inference in the \iid~case will necessitate many repeated evaluations of the individual likelihood $p_\mathrm{individual}(x_i | \theta)$.

\subsection{Traditional methods}
\label{sec:traditional}

The problem of inference without tractable likelihoods is not a new one, and two major approaches have been developed to address it. Arguably the most well-known is Approximate Bayesian Computation (\abc)~\citep{rubin1984, beaumont2002approximate}. Until recently, it was so established that the terms ``likelihood-free inference'' and ``\abc'' were often used interchangeably. In the simplest form of rejection \abc, the parameters $\theta$ are drawn from the prior,  the simulator is run with those values to sample $x_{\text{sim}} \sim p( \cdot | \theta)$, and $\theta$ is retained as posterior sample if the simulated data is sufficiently close to the observed data. In essence, the likelihood is approximated by the probability that the condition $\rho(x_{\text{sim}}, x_{\text{obs}}) < \epsilon$ is satisfied, where $\rho$ is some distance measure and $\epsilon$ is a tolerance. The accepted samples then follow an approximate version of the posterior. We show a schematic workflow of this algorithm in Fig.~\ref{fig:workflows}a (for a more elaborate Markov Chain Monte Carlo algorithm with a proposal function).

In the limit $\epsilon \to 0$, inference with \abc becomes exact, but for continuous data the acceptance probability vanishes. In practice, small values of $\epsilon$ require unfeasibly many simulations. For large $\epsilon$, sample efficiency is increased at the expense of inference quality. Similarly, the sample efficiency of \abc scales poorly to high-dimensional data $x$. Since the data immediately affect the rejection process (and in more advanced \abc algorithms the proposal distribution), inference for new observations requires repeating the entire inference algorithm. \abc is thus best-suited for the case of a single observation or at most a few \iid~data points.

Lacking space to do the vast \abc literature justice, we refer the reader to a review of \abc methods, see Ref.~\cite{sisson2018handbook}, and highlight the combination with MCMC~\citep{marjoram2003markov} and Sequential Monte Carlo (SMC)~\citep{sisson2007sequential,peters2012sequential}.

The second classical approach to simulation-based inference is based on creating a model for the likelihood by estimating the distribution of simulated data with histograms or kernel density estimation~\cite{Diggle1984MonteCM}. Frequentist and Bayesian inference then proceeds as if the likelihood were tractable. We sketch this algorithm in Fig.~\ref{fig:workflows}e (replacing the green learning step with a classical density estimation method). This approach has enough similarities to \abc to be dubbed ``Approximate Frequentist Computation'' by the authors of Ref.~\cite{Brehmer:2018eca}.  One advantage over \abc is that it is \emph{amortized}: after an upfront computational cost at the simulation and density estimation stage, new data points can be evaluated efficiently. In Fig.~\ref{fig:workflows}e this manifests itself as the blue ``data'' box only entering at the inference stage and not affecting the expensive simulation step. This property makes density estimation--based inference particularly well-suited for problems with many \iid~observations, a key reason for its wide-spread use in particle physics measurements.

Both of the traditional approaches suffer from the curse of dimensionality: in the worst case, the required number of simulations increases exponentially with the dimension of the data $x$. Therefore both approaches rely on low-dimensional summary statistics $y(x)$ and the quality of inference is tied to how well those summaries retain information about the parameters $\theta$. Traditionally, the development of powerful summary statistics has been the task of a domain expert, and the summary statistics have been prescribed prior to inference.

\section{Frontiers of simulation-based inference}
\label{sec:frontiers}

These traditional simulation-based inference techniques have played a key role in several fields for years. However, they suffer from shortcomings in three crucial aspects:
\begin{itemize}
    \item \emph{Sample efficiency}: Both \abc and classical density estimation techniques suffer from the curse of dimensionality. The poor scaling means that the number of simulated samples needed to provide a good estimate of the likelihood or posterior can be prohibitively expensive.
    \item \emph{Quality of inference}: The reduction of the data to low-dimensional summary statistics invariably discards some of the information in the data about $\theta$, which results in a loss in statistical power. Large values of the $\epsilon$ parameter in \abc or bandwidth parameter for kernel density estimation lead to poor approximations of the true likelihood. Both reduce the overall quality of inference.
    \item \emph{Amortization}: Performing inference with \abc for a new set of observed data requires repeating most steps of the inference chain, in particular if the proposal distribution depends on the observed data. The method scales poorly when applied to large numbers of observations. On the other hand, inference based on density estimation is amortized: the computationally expensive steps do not have to repeated for new observations. This is particularly desirable for the case with \iid~observations.
\end{itemize}

In recent years, new capabilities have become available that let us improve all three of these aspects. We loosely group them into three main directions of progress:
\begin{enumerate}
    \item The revolution in machine learning allows us to work with higher-dimensional data, which can improve the quality of inference. Inference methods based on neural network surrogates are directly benefitting from the impressive rate of progress in deep learning.
    \item Active learning methods can systematically improve sample efficiency, letting us tackle more computationally expensive simulators.
    \item The deep integration of automatic differentiation and probabilistic programming into the simulation code, as well as the augmentation of training data with additional information that can be extracted from the simulator, are changing the way the simulator is treated in inference: it is no longer a black box, but exposed to the inference workflow.
\end{enumerate}

\begin{figure}
    \centering
    \includegraphics[width=0.4\textwidth]{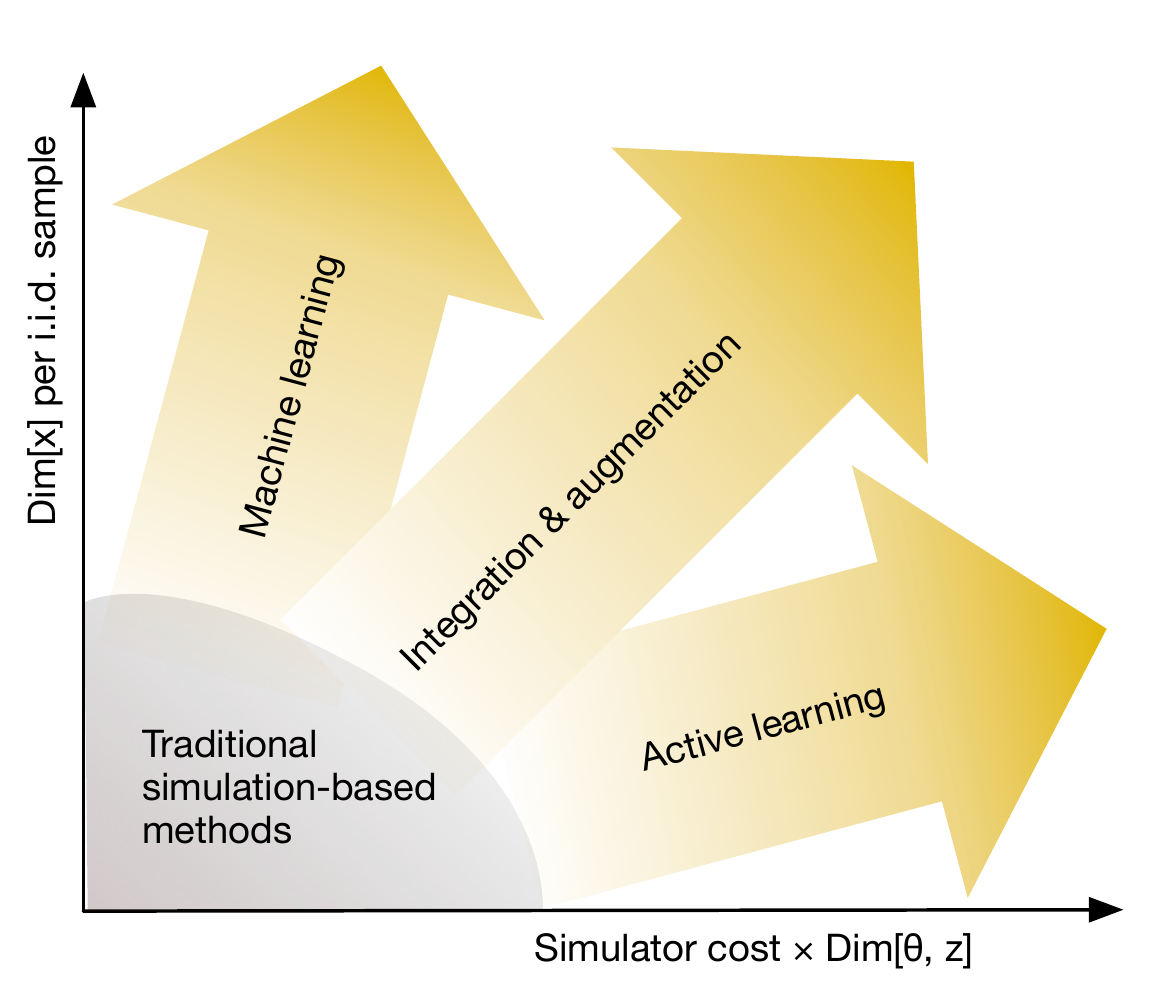}
    \caption{A schematic illustration of how machine learning, active learning, and  integration of automatic differentiation and probabilistic programming into the simulation code  are expanding the frontier of traditional approaches to simulation-based inference. }
    \label{fig:frontiers}
\end{figure}

We sketch these trends in Fig.~\ref{fig:frontiers}, broadly categorizing the landscape of inference tasks in a two-dimensional plane of the dimensionality of the data (vertical axis) and the complexity of the simulator (horizontal axis).

\subsection{A revolution in machine learning}
\label{sec:tools}

Over the last decade, machine learning techniques, in particular deep neural networks, have turned into versatile, powerful, and popular tools for a variety of problems~\cite{2015Natur.521..436L}. Neural networks initially demonstrated breakthroughs in supervised learning tasks such as classification and regression. They can easily be composed to solve higher-level tasks, lending themselves to problems with a hierarchical or compositional structure. Architectures tailored to various data structures have been developed, including dense or fully connected networks aimed at unstructured data, convolutional neural networks that leverage spatial structure for instance in image data, recurrent neural networks for variable-length sequences, and graph neural networks for graph-structured data. Choosing an architecture well suited for a specific data structure is an example of \emph{inductive bias}, which more generally refers to the assumptions inherent in a learning algorithm independent of the data. Inductive bias is one of the key ingredients behind most successful applications of deep learning, though it is difficult to characterize its role precisely.

One area where neural networks are being actively developed is density estimation in high dimensions: given a set of points $\{x\} \sim p(x)$, the goal is to estimate the probability density $p(x)$. As there are no explicit labels, this is usually considered an unsupervised learning task. We have already discussed that classical methods based for instance on histograms or kernel density estimation do not scale well to high-dimensional data. In this regime, density estimation techniques based on neural networks are becoming more and more popular. One class of these neural density estimation techniques are \emph{normalizing flows}~\cite{2014arXiv1410.8516D, 2016arXiv160508803D, 2018arXiv180703039K, 2015arXiv150203509G, 2016arXiv160502226U, 2016arXiv160106759V, 2016arXiv160605328V, 2016arXiv160903499V, 2016arXiv160604934K, 2017arXiv170507057P, 2018arXiv180400779H, 2019arXiv190805164W, 2019arXiv190602145D, 2019arXiv190604032D, DBLP:journals/corr/abs-1806-07366, 2018arXiv181001367G}, in which variables described by a simple base distribution $p(u)$ such as a multivariate Gaussian are transformed through a parameterized invertible transformation $x = g_\phi(u)$ that has a tractable Jacobian. The target density $p_g(x)$ is then given by the change-of-variables formula as a product of the base density and the determinant of the transformation's Jacobian. Several such steps can be stacked, with the probability density ``flowing'' through the successive variable transformations. The parameters $\phi$ of the transformations are trained by maximizing the likelihood of the observed data under the model $p_g(x_\mathrm{obs})$, resulting in a model density that approximates the true, unknown density $p(x)$. In addition to having a tractable density, it is possible to generate data from the model by drawing the hidden variables $u$ from the base distribution and applying the flow transformations. Neural density estimators have been generalized to model the dependency on additional inputs, \ie to model a conditional density such as the likelihood $p(x|\theta)$ or posterior $p(\theta|x)$.

Another class of approaches use autoregressive models, in which the probability distribution of a high-dimensional variable is factorized into successive conditional densities of the individual components~\citep{2015arXiv150203509G, 2016arXiv160502226U, 2016arXiv160106759V, 2016arXiv160605328V, 2016arXiv160903499V, 2016arXiv160604934K, 2017arXiv170507057P, 2018arXiv180400779H, 2019arXiv190805164W}. These models are expressive, have a tractable (conditional) density, and can be used to generate synthetic data. While autoregressive models are somewhat disfavored in industrial applications because generating samples from them can be slow, the sequential nature is more closely aligned with the way simulators are written and offers an opportunity to align the neural networks latent variables with the semantically meaningful latent variables of simulators.

{Generative Adversarial Networks} (GANs) are an alternative type of generative model based on neural networks. Unlike in normalizing flows and autoregressive models, the transformation implemented by the generator is not restricted to be invertible. While this allows for more expressiveness, the density defined by the generator is intractable. Since maximum likelihood is not a possible training objective, the generator is pitted against an adversary, whose role is to distinguish the generated data from the target distribution. We will later discuss how the same idea can be used for simulation-based inference, using an idea known as the ``likelihood ratio trick''.

\subsection{Active learning}

A simple, but very impactful idea is to run the simulator at parameter points that are expected to increase our knowledge the most. This can be done iteratively such that after each simulation the knowledge resulting from all previous runs is used to guide which parameter point should be used next. There are multiple technical realizations of this idea of active learning. It is commonly applied in a Bayesian setting, where the posterior can be continuously updated and used to steer the proposal distribution of simulator parameters~\cite{2014arXiv1401.2838M, 2015arXiv150103291G, 2015arXiv150603693M, 2017arXiv170400520J, 2017arXiv170309930W, 2017arXiv171101861L, 2018arXiv180507226P}. But it applies equally well to efficiently calculating frequentist confidence sets~\cite{ranjan2008sequential,Bect2012,lukas_heinrich_2018_1634428}. Even simple implementations can lead to a substantial improvement in sample efficiency.

Similar ideas are discussed in the context of decision making, experimental design, and reinforcement learning, and we expect further improvements in inference algorithms from the cross-pollination between these fields. For instance, a question that is occasionally discussed in the context of reinforcement learning~\citep{cutler2014reinforcement, hamrick2017metacontrol} or Bayesian optimization~\cite{kandasamy2017multi}, but has not yet been applied to the likelihood-free setting, is how to make use of multi-fidelity simulators offering multiple levels of precision or approximations.

\subsection{Integration and augmentation}
\label{sec:integration}

Both machine learning and active learning can substantially improve quality of inference and sample efficiency compared to classical methods. However, overall they do not change the basic approach to simulation-based inference dramatically: they still treat the simulator as a generative black box that takes parameters as input and provides data as output, with a clear separation between the simulator and the inference engine. A third direction of research is changing this perspective, opening this black box to access more information and integrating inference and simulation more tightly.

One example of this shift is the probabilistic programming paradigm. Gordon~et~al.~\cite{Gordon2014ProbabilisticP} describe probabilistic programs as the usual functional or imperative programs with two added constructs: (1) the ability to draw values at random from distributions, and (2) the ability to condition values of variables in a program via observations. We have already described simulators as probabilistic programs focusing on the first construct, which does not require opening the black box. However, conditioning on the observations requires a deeper integration as it involves controlling the randomness in the generative process. This approach abstracts the capabilities needed to implement Particle Filters and SMC~\cite{doucet2009tutorial}.  Previously, this required writing the program in a special purpose language; however, recent work allows these capabilities to be added to existing simulators with minimal changes to their codebase~\cite{2019arXiv190703382G}. Ultimately, probabilistic programming aims at providing the tools to infer the incredibly complex space of all execution traces of the simulator conditioned on the observation.

A complementary development is the observation that additional information that characterizes the latent data-generating process can be extracted from the simulator and used to augment the data used to train surrogates. This augmented training data can be exploited in supervised learning objectives, and can dramatically increase the sample efficiency for surrogate training. Those developing inference algorithms and those familiar with the details of the simulator should consider whether, in addition to the sole ability to sample $x \sim p(x|\theta)$, the following properties in which the corresponding quantities in the simulator are well-defined and tractable.
\begin{enumerate}[label=\Roman*.]
    \item $p(x | z, \theta)$: the probability density of the output data given the latent variables.
    \item $t(x,z|\theta) \equiv \nabla_\theta \log p(x, z | \theta)$: the joint score is the gradient of the joint log probability density of output data and latent variables with respect to the parameters.
    \item $\nabla_z \log p(x, z | \theta)$: the gradient of the same quantity, but with respect to the latent variables.
    \item  $r(x,z| \theta, \theta') \equiv p(x, z | \theta) / p(x, z | \theta’)$: the ratio of the joint probability density of output data and latent variables for two parameter points $\theta$ and $\theta'$.
    \item $\nabla_\theta (x,z)$: the derivative of output data and latent variables with respect to the parameters.
    \item $\nabla_z x$: the gradient of the output data with respect to the latent variables.
\end{enumerate}
These quantities can then be used to augment the usual output $x$ from the simulator and can be exploited in supervised learning objectives, and can dramatically increase the sample efficiency for surrogate training~\cite{graham2017asymptotically, Brehmer:2018hga, Stoye:2018ovl}, as we will detail later.

Many of the quantities above involve derivatives, which can now be efficiently calculated using automatic differentiation (often referred to simply as \textit{autodiff})~\cite{baydin2018automatic}.
Autodiff is a family of techniques similar to but more general than the backpropagation algorithm that is ubiquitous in deep learning. Automatic differentiation, like probabilistic programming, involves non-standard interpretations of the simulation code and has been developed by a small but established field of computer science.  In recent years several have advocated that deep learning would be better described as \textit{differential programming}~\cite{olah_2015,lecun_2018}. With this view, incorporating autodiff into existing simulation codes is a more direct way to exploit the advances in deep learning than trying to incorporate domain knowledge into an entirely foreign substrate such as a deep neural network.

Extracting the necessary information from the simulator again requires integration deep in the code.  While technologies to incorporate probabilistic programming paradigm into existing code bases are just emerging, the development of tools to enable autodiff in the most commonly used scientific programming languages is well advanced. We highlight that two of the quantities above (II and III) involve both autodiff and probabilistic programming. The integration of inference and simulation as well as the idea of augmenting the training data with additional quantities have the potential to change the way we think about simulation-based inference. In particular, this perspective can influence the way simulation codes are developed in order to provide these new capabilities.

\section{Workflows for simulation-based Inference}
\label{sec:inference}

This wide array of capabilities can be combined in different inference workflows. Some of these are structurally identical to the traditional \abc and density estimation--based methods, while others are fundamentally different. As a guideline through this array of different workflows, let us first discuss common building blocks, and the different approaches that can be taken in each of these components. In Fig.~\ref{fig:workflows} and the following sections we will then piece these blocks together into different inference algorithms.

An integral part of all inference methods is running the simulator, visualized as a yellow pentagon in Fig.~\ref{fig:workflows}. The parameters at which the simulator is run are drawn from some proposal distribution, which may or may not depend on the prior in a Bayesian setting, and can either be chosen statically, or iteratively with an active learning method. Next, the potentially high-dimensional output from the simulator may be used directly as input to the inference method, or reduced to low-dimensional summary statistics, which may be prescribed or learned from data.

The inference techniques can be broadly separated into those which, like \abc, use the simulator itself during inference, and methods which construct a surrogate model and use that for inference. In the first case, the output of the simulator is directly compared to data, see the top panels of Fig.~\ref{fig:workflows}. In the latter case, the output of the simulator is used as training data for an estimation or machine learning stage, shown as the green boxes in the bottom panels of Fig.~\ref{fig:workflows}. The resulting surrogate models, shown as red hexagons, are then used for inference.

The algorithms address the intractability of the true likelihood in different ways: some methods construct a tractable surrogate for the likelihood function, others for the likelihood \emph{ratio} function, both of which make frequentist inference straightforward. In other methods, the likelihood function never appears explicitly, for instance when it is implicitly replaced by rejection probability (an approach that does not lend itself to frequentist inference).

The final target for Bayesian inference is the posterior. Methods differ in whether they provide access to samples of parameter points sampled from the posterior, for instance from MCMC or ABC, or a tractable function that approximates the posterior function. Similarly, some methods require specifying which quantities are to be inferred early on in the workflow, while others allow this decision to be postponed.

\subsection{Using the simulator directly during inference}

\begin{figure*}[!t]
    \centering%
    \includegraphics[width=\textwidth]{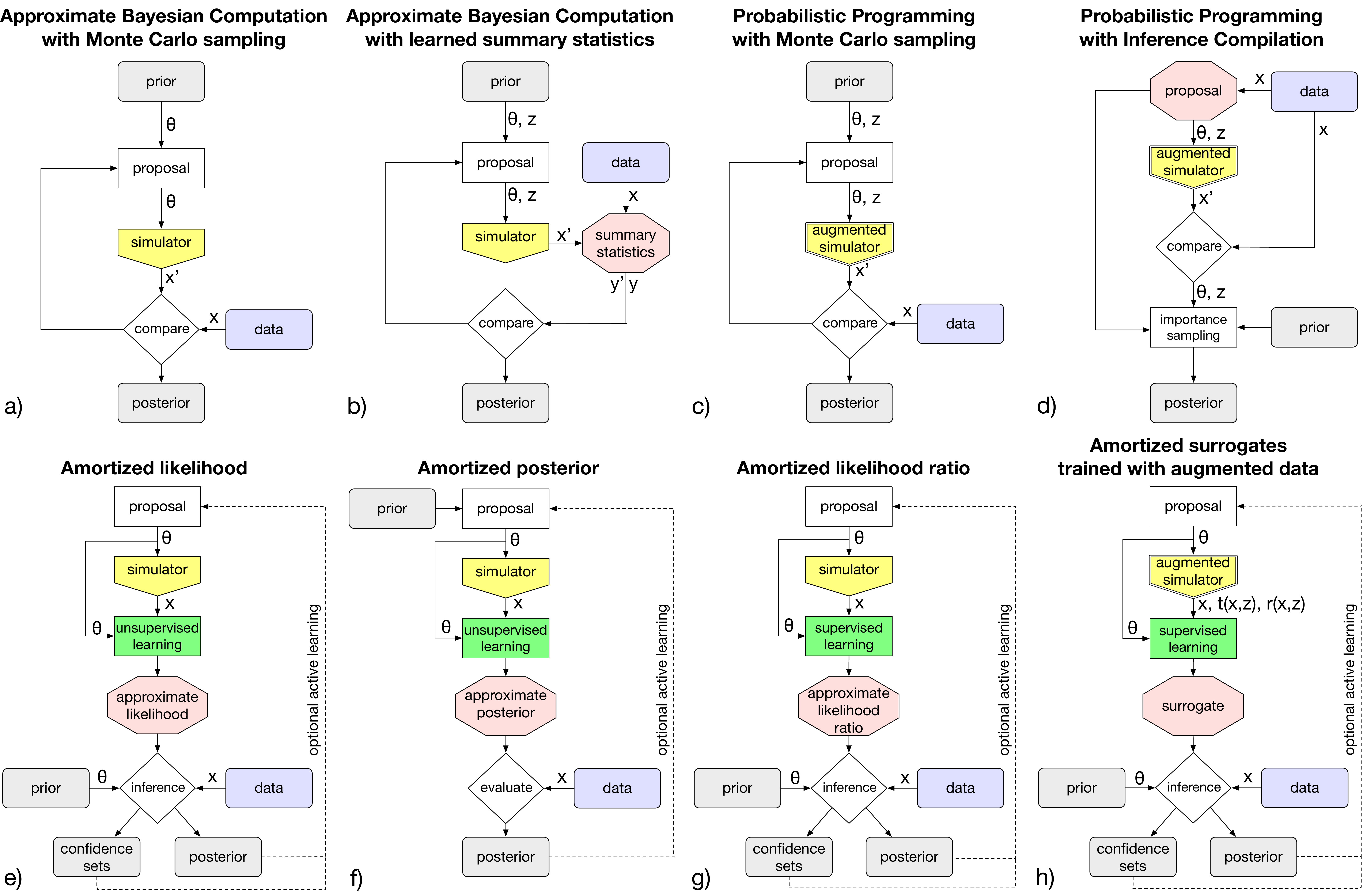}%
    \caption{Overview of different approaches to simulation-based inference.}%
    \label{fig:workflows}%
\end{figure*}

Let us now discuss how these blocks and computational capabilities can be combined into inference techniques, beginning with those which, like \abc, use the simulator directly during inference. We sketch some of these algorithms in the top panels of Fig.~\ref{fig:workflows}.

One of the major shortcomings of \abc is its reliance on low-dimensional summary statistics. Classifier \abc~\cite{gutmann2017likelihood} removes the requirement of compressing the data into summary statistics by instead training a classifier to estimate the discrepancy between observed and simulated data.

A reason for the poor sample efficiency of the original rejection \abc algorithm is that the simulator is run at parameter points drawn from the prior, which may have a large mass in regions that are in strong disagreement with the observed data. Different algorithms have been proposed that instead run the simulator at parameter points that are expected to improve the knowledge on the posterior the most~\cite{2014arXiv1401.2838M, 2015arXiv150103291G, 2015arXiv150603693M, 2017arXiv170400520J, 2017arXiv170309930W}. Compared to vanilla \abc, these techniques improve sample efficiency, though they still require the choice of summary statistics, distance measure $\rho$, and tolerance $\epsilon$.

In the case where the final stage of the simulator is tractable or the simulator is differentiable (respectively, properties I and VI from the list in Sec.~\ref{sec:frontiers}.\ref{sec:integration}), asymptotically exact Bayesian inference is possible~\cite{graham2017asymptotically} without relying on a distance tolerance or summary statistics, removing \abc's main limitations in terms of quality of inference.

The probabilistic programming paradigm presents a more fundamental change to how inference is performed. First, it requires the simulator to be written in a probabilistic programming language, though recent work allows these capabilities to be added to existing simulators with minimal changes to their codebase~\cite{2019arXiv190703382G}. In addition, probabilistic programming either requires a tractable likelihood for the final step $p(x|z, \theta)$ (quantity I) or the introduction of an \abc-like comparison. When these criteria are satisfied, several inference algorithms exist that can draw samples from the posterior $p(\theta, z | x)$ of the input parameters $\theta$ and the latent variables $z$ given some observed data $x$. These techniques are either based on MCMC, see Fig.~\ref{fig:workflows}c, or on training a neural network to provide proposal distributions~\cite{le2017inference} as shown in Fig.~\ref{fig:workflows}d. The key difference to \abc is that the inference engine controls all steps in the program execution and can bias each draw of random latent variables to make the simulation more likely to match the observed data, improving sample efficiency.

A strength of these algorithms is that they allow to infer not only the input parameters into the simulator, but the entire latent process leading to a particular observation. This allows us to answer entirely different questions about scientific processes, adding a particular kind of physical interpretability that methods based on surrogates do not possess. While standard \abc algorithms in principle allow for inference on $z$, probabilistic programming solves this task more efficiently.

\subsection{Surrogate models}

A key disadvantage of using the simulator directly during inference is the lack of amortization. When new observed data becomes available, the whole inference chain has to be repeated. By training a tractable surrogate or emulator for the simulator, inference is amortized: after a (computationally expensive) upfront simulation and training phase, new data can be evaluated very efficiently. This approach scales particularly well to data consisting of many \iid~observations. As discussed in Sec.~\ref{sec:sbi}\ref{sec:traditional}, this is not a new idea, and well-established methods use classical density estimation techniques to create a surrogate model for the likelihood function. But the new computational capabilities discussed in Sec.~\ref{sec:tools} have given new momentum to this class of inference techniques. They can be incorporated in various ways, providing approximations of the true parameters given data, defining suitable summary statistics, learning the likelihood, the likelihood ratio, or the posterior; we will briefly go through these different options to organize the inference. Selected algorithms are visualized in the bottom panels of Fig.~\ref{fig:workflows}.

Perhaps the most obvious of these approaches is to directly invert the parameter-to-data process implemented by the simulator and to train a model to estimate the true parameters $\hat{\theta}(x)$ as a function of observed data $x$~\citep{tarantola2005inverse, 2017arXiv170604008P, Pesah:2018tbc, 2017arXiv170707113L}. However, point estimates are not always useful and a probabilistic interpretation in terms of the likelihood or posterior of these methods is not obvious, so will not focus on this approach.

A powerful probabilistic approach is to train neural conditional density estimators such as normalizing flows as a surrogate for the simulator. The conditional density can be defined in two directions: the network can either learn the posterior $p(\theta|x)$~\citep{2015arXiv150505770J, NIPS2016_6084, 2017arXiv171101861L, 2018arXiv180505480I, 2016arXiv160206701P, 2017arXiv170208896T} or the likelihood $p(x|\theta)$~\cite{2018arXiv180507226P, 2018arXiv181108723D, 2018arXiv180509294L, Alsing:2019xrx}. We show these three techniques in Figs.~\ref{fig:workflows}e and \ref{fig:workflows}f; note that the likelihood surrogate algorithm is structurally identical to the classical density estimation--based approach, but uses more powerful density estimation techniques.

Relatedly, neural networks can be trained to learn the likelihood ratio function $p(x|\theta_0) / p(x|\theta_1)$ or $p(x|\theta_0) / p(x)$, where in the latter case the denominator is given by a marginal model integrated over a proposal or the prior~\cite{Neal:2007zz, 2012arXiv1212.1479F, Cranmer:2015bka, 2016arXiv161003483M, 2016arXiv161110242D, gutmann2017likelihood, 2018arXiv181009899D, Hermans:2019ioj, Andreassen:2019nnm}. We sketch this approach in Fig.~\ref{fig:workflows}g. The key idea is closely related to the discriminator network in GANs mentioned above: a classifier is trained using supervised learning to discriminate two sets of data, though in this case both sets come from the simulator and are generated for different parameter points $\theta_0$ and $\theta_1$. The classifier output function can be converted into an approximation of the likelihood ratio between $\theta_0$ and $\theta_1$! This This manifestation of the Neyman-Pearson lemma in a machine learning setting is often called the \emph{likelihood ratio trick}.

These three surrogate-based approaches are all amortized: after an upfront simulation and training phase, the surrogates can be evaluated efficiently for arbitrary data and parameter points. They require an upfront specification of the parameters of interest, the network then implicitly marginalizes over all other (latent) variables in the simulator. All three classes of algorithms can employ active learning elements such as an iteratively updated proposal distribution to guide the simulator parameters $\theta$ towards relevant parameter region, improving sample efficiency. Using neural networks eliminates the requirement of low-dimensional summary statistics, leaving it to the employed model to learn the structures  in high-dimensional data and potentially improving quality of inference.

Despite these fundamental similarities, there are some differences between emulating the likelihood, the likelihood ratio, and the posterior. Learning the posterior directly provides the main target quantity in Bayesian inference, but induces a prior dependence at every stage of the inference method. Learning the likelihood or the likelihood ratio enables frequentist inference or model comparisons, though for Bayesian inference an additional MCMC or VI step is necessary to generate samples from the posterior. The prior independence of likelihood or likelihood ratio estimators also leads to extra flexibility to change the prior during inference. An advantage of training a generative model to approximate the likelihood or posterior over learning the likelihood ratio function is the added functionality of being able to sample from the surrogate model. On the other hand, learning the likelihood or posterior is an unsupervised learning problem, whereas estimating the likelihood ratio through a classifier is an example of supervised learning and often a simpler task. Since for the higher-level inference goal the likelihood and the likelihood ratio can be used interchangeably, learning a surrogate for the likelihood ratio function may often be more efficient.

Another strategy that allows us to leverage supervised learning is based on extracting additional quantities  from the simulator that characterize the likelihood of the latent process (\eg  II and IV from the list in Sec.~\ref{sec:frontiers}.\ref{sec:integration}). This additional information can be used to augment the training data for surrogate models. The resulting supervised learning task can often be solved more efficiently, ultimately improving the sample efficiency in the inference task~\cite{Brehmer:2018hga, Brehmer:2018kdj, Brehmer:2018eca, Stoye:2018ovl}.

Surrogate-based approaches benefit from imposing suitable inductive bias for a given problem. It is widely acknowledged that the network architecture of a neural surrogate should be chosen according to the data structure (\eg images, sequences, or graphs). Another, potentially more consequential, way of imposing inductive bias is to have the surrogate model reflect the causal structure of the simulator. Manually identifying the relevant structures and designing appropriate surrogate architectures is very domain-specific, though has been shown to improve the performance on some problems~\cite{Louppe:2017ipp, Andreassen:2018apy, Carleo:2019ptp}. Recently attempts are being made to automate the process of creating surrogates that mimic the simulation~\cite{munk2019deep}. Looking further ahead, one would like to learn surrogates that reflect the causal structure of a coarse grained system. If this is possible, it would allow the surrogate to model only the relevant degrees of freedom for the phenomena that emerge from the underlying mechanistic model.

\subsection{Preprocessing and postprocessing}

There are a number of additional steps that can surround these core inference algorithms, either in the form of preprocessing steps that precede the main inference stage, or as an afterburner following the main inference step.

One preprocessing step is to learn powerful summaries $y(x)$. Because of the curse of dimensionality, both \abc and classical density estimation--based inference methods require a compression of the data into low-dimensional summary statistics. They are usually prescribed, \ie hand-chosen by domain scientists based on their intuition and knowledge of the problem at hand, but the resulting summaries will generally lose some information compared to the original data. A minimally invasive extension of these algorithms is to first learn summary statistics that have certain optimality properties, before running a standard inference algorithm such as \abc. We sketch this approach in Fig.~\ref{fig:workflows}b for \abc, but it applies equally to inference based on density estimation.

The \emph{score} $t(x|\theta) \equiv \nabla_\theta p(x|\theta)$, the gradient of the log (marginal) likelihood with respect to the parameters of interest, defines such a vector of optimal summary statistics: in a neighborhood of $\theta$, the score components are sufficient statistics, and they can be used for inference without loss of information. Just like the likelihood, the score itself is generally intractable, but it can be estimated based on quantity V and an exponential family approximation~\cite{Alsing:2017var, Alsing:2018eau}. If quantity II is available, augmented data extracted from the simulator can instead be used to train a neural network to estimate the score~\cite{Brehmer:2018hga} without requiring such an approximation. Learned summaries can also be made robust with respect to nuisance parameters~\cite{deCastro:2018mgh, Alsing:2019dvb}.

Even if it is not necessary to reduce the data to low-dimensional summary statistics, in some fields the measured raw or ``low-level'' data can be very high-dimensional. It is then common practice to compress them to a more managable set of ``high-level'' features of moderate dimensionality and to use these compressed data as input to the inference workflow.

Inference compilation~\cite{le2017inference} is a preprocessing step for probabilistic programming algorithms, shown in Fig.~\ref{fig:workflows}d. Initial runs of the simulator are used to train a neural network used for sequential importance sampling of both the parameters $\theta$ and the latent variables $z$.

After the completion of the core inference workflow, an important question is whether the results are reliable: can the outcome be trusted in the presence of imperfections such as limited sample size, insufficient network capacity, or inefficient optimization?

One solution is to calibrate the inference results. Using the ability of the simulator to generate data for any parameter point, we can use a parametric bootstrap approach to calculate the distribution of any quantity involved in the inference workflow. These distributions can be used to calibrate the inference procedure to provide confidence sets and posteriors with proper coverage and credibility~\citep{Cranmer:2015bka, Brehmer:2018eca}. While possible in principle, such procedures may require a large number of simulations.

Other diagnostic tools that can be applied at the end of the inference stage involve training classifiers to distinguish data from the surrogate model and the true simulator~\citep{Cranmer:2015bka}, checking known expectation values of estimators of the likelihood, likelihood ratio, or score~\citep{Brehmer:2018eca}; varying reference parameters that should leave the inference result invariant~\citep{Cranmer:2015bka}; ensemble methods; and comparing distributions of network output against known asymptotic properties~\citep{Wilks:1938dza, Wald, Cowan:2010js}. Passing these sanity checks does not guarantee that an estimator is correct, but failing them is an indication of problems. Some of these methods may be used for uncertainty estimation, though the statistical interpretation of such error bars is not always obvious.

None of these diagnostics address the issues encountered if the model is misspecified and the simulator is not an accurate description of the system being studied. Model misspecification is a problem that plagues inference with both prescribed and implicit models equally. Usually this is addressed by expanding the model to have more flexibility and introducing additional nuisance parameters. 

\subsection{Recommendations}

The considerations needed to choose which of the approaches described above is best for a given problem will include the inference goals, the dimensionality of model parameters, the latent variables, and the data; whether good summary statistics are available; the internal structure of the simulator; the computational cost of the simulator; the level of control over how the simulator is run; and on whether the simulator is a black box or whether any of the quantities discussed in Sec.~\ref{sec:frontiers}.\ref{sec:integration} can be extracted from it. Nevertheless, we believe that the existing body of research lets us provide a few general guidelines.

First, if any of the quantities discussed in Sec.~\ref{sec:frontiers} are available, they should be leveraged. Powerful algorithms are available for the case with differentiable simulators~\cite{graham2017asymptotically}, for simulators for which the joint likelihood of data and latent variables is accessible~\cite{Brehmer:2018hga}, and for simulators explicitly written as a probabilistic model in a probabilistic programming framework~\cite{wood-aistats-2014}. Probabilistic programming is also the most versatile approach when the goal is not only inference on the parameters $\theta$, but also the latent variables $z$. 

If powerful low-dimensional summary statistics are established, traditional techniques can still offer a reasonable performance. In most cases, however, we recommend trying methods based on training a neural network surrogate for the likelihood~\cite{2018arXiv180507226P, 2018arXiv180509294L} or the likelihood ratio~\cite{Cranmer:2015bka, gutmann2017likelihood, Hermans:2019ioj}. If generating synthetic data from the surrogate is not important, learning the likelihood ratio  rather than the likelihood allows us to leverage powerful supervised learning methods.

Finally, active learning techniques can improve the sample efficiency for all inference techniques. There is a tradeoff between active learning, which tailors the efficiency to a particular observed data set, and amortization, which benefits from  surrogates that are agnostic about the observed data. A good compromise here will  depend on the number of  observations and the sharpness of the posterior compared to the prior.

\section{Discussion}
\label{sec:conclusions}

Until recently, scientists confronted with inverse problems and a complex simulator as a forward model had little recourse than to choose \abc or methods based on classical density estimation techniques. While these approaches have served some domains of science quite well, they have relied heavily on experts providing powerful summary statistics. As a result, the techniques are labor intensive and do not lend themselves well to high-dimensional data where powerful summary statistics are not obvious. While not explicit, there is a frontier where these traditional methods are no longer useful, beyond which scientists must resort to other heuristics not framed as a statistical statement tied to the underlying mechanistic model.

The term likelihood-free inference has served as a point of convergence for what were previously disparate communities, and a new lingua franca has emerged. This has catalyzed significant cross-pollination and led to a renaissance in simulation-based inference. The advent of powerful machine learning methods is enabling practitioners to work directly with high dimensional data and to reduce the reliance on expert-crafted summary statistics. New programming paradigms such as probabilistic programming and differentiable programming provide new capabilities that enable entirely new approaches to simulation-based inference. Finally, taking a more systems-level view of simulation-based inference that brings together the statistical and computational considerations has taken root. Here active learning is leading the way, but we expect more advances like this as simulation-based inference matures.

The rapidly advancing frontier means that several domains of science should expect either a significant improvement in inference quality or the transition from heuristic approaches to those grounded in statistical terms tied to the underlying mechanistic model. It is not unreasonable to expect that this transition may  have a profound impact on science.


\acknow{KC and JB are supported by the National Science Foundation under the awards ACI-1450310, OAC-1836650, and OAC-1841471 and by the Moore-Sloan data science environment at NYU. GL is recipient of the ULi\`{e}ge-NRB Chair on Big Data and is thankful for the support of NRB.}
\showacknow{}

\bibliography{references}

\end{document}